\documentclass[conference,letterpaper]{IEEEtran}

\usepackage[utf8]{inputenc} 
\usepackage[T1]{fontenc}
\usepackage{url}
\usepackage{ifthen}
\usepackage{cite}
\usepackage[cmex10]{amsmath}
\usepackage{xcolor}
\usepackage{bm}
\usepackage{amssymb}
\usepackage{multirow}
\usepackage{hyperref}

\usepackage{graphicx}
\usepackage[labelfont=bf]{caption}
\usepackage{subfigure}
\usepackage{flushend}

\definecolor{DarkGreen}{rgb}{0.1,0.5,0.1}
\definecolor{DarkRed}{rgb}{0.5,0.1,0.1}
\definecolor{DarkBlue}{rgb}{0.1,0.1,0.5}
\definecolor{DarkPurple}{rgb}{0.5,0.2,0.5}
\definecolor{DarkTurquoise}{rgb}{0.1,0.5,0.5}

\begin{document}
\title{TexShape: Information Theoretic Sentence Embedding for Language Models\vspace{-0.1cm}}

\author{
\IEEEauthorblockN{Kaan Kale$^{*{1}}$, Homa Esfahanizadeh$^{*{2}}$, Noel Elias$^{3}$, Oguzhan Baser$^{3}$, Muriel M\'edard$^{4}$, and Sriram Vishwanath$^{3}$}
\IEEEauthorblockA{
$^{1}$Boğaziçi University, Istanbul, Türkiye, Email: huseyin.kale@std.bogazici.edu.tr
}
\IEEEauthorblockA{
$^{2}$Nokia Bell Labs, Murray Hill, NJ, USA, Email: homa.esfahanizadeh@nokia-bell-labs.com
}
\IEEEauthorblockA{
$^{3}$University of Texas at Austin, Austin, TX, USA, Emails: \{nelias,oguzhanbaser,sriram\}@utexas.edu
}
\IEEEauthorblockA{
$^{4}$Massachusetts Institute of Technology (MIT), Cambridge, USA, Email: medard@mit.edu
}
\IEEEauthorblockA{${^{*}}$These authors contributed equally to this work.}\vspace{-0.9cm}
}

\maketitle


\begin{abstract}
   With the exponential growth in data volume and the emergence of data-intensive applications, particularly in the field of machine learning, concerns related to resource utilization, privacy, and fairness have become paramount. This paper focuses on the textual domain of data and addresses challenges regarding encoding sentences to their optimized representations through the lens of information-theory. In particular, we use empirical estimates of mutual information, using the Donsker-Varadhan definition of Kullback–Leibler divergence. Our approach leverages this estimation to train an information-theoretic sentence embedding, called TexShape, for (task-based) data compression or for filtering out sensitive information, enhancing privacy and fairness. In this study, we employ a benchmark language model for initial text representation, complemented by neural networks for information-theoretic compression and mutual information estimations. Our experiments demonstrate significant advancements in preserving maximal targeted information and minimal sensitive information over adverse compression ratios, in terms of predictive accuracy of downstream models that are trained using the compressed data.
\end{abstract}

\section{Introduction}
The domain of machine learning (ML), particularly that of transformers and language models, has witnessed explosive growth over the past few years, e.g. \cite{devlin2018bert,NIPS2017_3f5ee243,radford2018improving} among many others. Specifically, the advancement of technologies such as ChatGPT \cite{radford2018improving}, Bard \cite{chowdhery2023palm}, LLaMa \cite{touvron2023llama}, etc. have produced significant strides within the field of natural language processing.
Essentially, the ability to extract representations from massive datasets and use such representations to classify and generate synthetic versions have found a multitude of applications, ranging from consumer engagement to enterprise productivity and efficiency \cite{zhao2023survey}. This process of transforming massive datasets into (generative) models is very promising and powerful \cite{nalisnick2018deep}. However, its connection with theoretical aspects of statistics and machine learning are currently poorly understood. In particular, we understand very little about the connections between information-theoretic quantities and concepts within (generative) ML. 

Since its inception \cite{Shannon48,el2011network}, information theory has formed the basis on which all data processing systems, including communication, signal processing, and compression, have been benchmarked and characterized \cite{Goldsmith_2005,OppenheimSignal}. Information theory has been used to understand both lossless and lossy compression of data as well as optimal mechanisms of (universal) compression of data sources. Indeed, methods to combine such information-theoretic understanding with representations in machine learning would be invaluable to improve the complexity and performance of ML systems. 

This paper is structured to be a very early step in this direction for large language models (LLMs). In particular, our goal is to develop optimized and succinct text representations for ML through the use of information-theoretic tools. We built upon the encoding steps in LLMs, where the utilization of information theory can result in improved and compressed data representations. This combination forms a perfect union as information theory provides insights into compression and representation of data, while ML models generally desire compressed representations for improved performance at lower complexities. Ultimately, such a connection can potentially result in the design of an ``optimal" encoder that has the best compressed representation for a desired level of performance. Furthermore, information-theoretic tools can also be employed to succinctly and measurably capture privacy and fairness constraints within an ML setting. 

Mutual information (MI) is an integral concept in information theory. In particular, MI is used to quantify the average related information between the outcomes of two random distributions \cite{Shannon48}, making it a great candidate for (task-oriented) sample representations \cite{Medard:IBprivfunnel}. 
However, one major drawback is that computing MI requires knowing the probability distributions that are often intractable to estimate in practice. A practical approach, known as variational information maximization \cite{IMalgoNIPS}, involves approximating the MI by maximizing the likelihood of estimating one random variable given the other, applied, for example, in generative applications \cite{NIPS2016_7c9d0b1f} and text compression \cite{cheng-etal-2020-improving}. Simpler and more accurate estimation of MI has also been pursued, recent work attracting much attention\cite{Belghazi:mine,Choi:remine,Song:smile,kim2023cryptomine}. These estimates, although very promising, still face practical challenges with respect to stability, precision, and complexity. Recently, a framework, InfoShape \cite{Esfahanizadeh_2023}, was presented to incorporate MI in the design of privacy-preserving image representations. {At a high level, our work extends the foundation laid by InfoShape, encompassing additional data modalities and tackling a broader spectrum of data processing tasks beyond the privacy-utility trade-off, such as fairness and semantic compression.}

Our contributions can be summarized as follows:
\begin{itemize}
\item We present an information-theoretic architecture for {text processing}, based on bidirectional encoder representations from transformers (BERTs) \cite{devlin2018bert}, deep neural networks (DNNs), and variational MI inference \cite{Donsker:dvkl,kraskov2004estimating}. 
\item We utilize these information-theoretical {tools} to enable custom (task-based) text compression to enhance desired features within the ML system, such as better resource management, improved privacy, and/or greater fairness.
\end{itemize}
{We have provided public access to our code and data at \href{https://github.com/hesfahanizadeh/NeuralInformationShaping/tree/main/TexShape}{https://github.com/hesfahanizadeh/NeuralInformationShaping}.}

\section{Preliminaries and Problem Statement}
In this paper, scalars and matrices are represented with lowercase and uppercase letters, respectively. Random variables are distinguished from deterministic ones using the boldface style.

\subsection{Preliminaries}

Consider two random variables $\bm{X}\in\mathcal{X}$ and $\bm{Z}\in\mathcal{Z}$, where $\mathcal{X}$ and $\mathcal{Z}$ indicate the space set of the two random variables. The MI value $\mathcal{I}(\bm{X};\bm{Z})$ quantifies how much information is obtained on average about $\bm{X}$ by observing $\bm{Z}$. By Shannon's definition \cite{Shannon48},\vspace{-0.2cm}
\begin{equation*}
    \mathcal{I}(\bm{X};\bm{Z})=\sum_{X\in\mathcal{X}}\sum_{Z\in\mathcal{Z}}P_{\bm{X,Z}}(X,Z)\log(\frac{P_{\bm{X,Z}}(X,Z)}{P_{\bm{X}}(X)P_{\bm{Z}}(Z)}),
\end{equation*}
where $P_{\bm{X,Z}}(X,Z)$ is the joint probability mass function (PMF) of $\bm{X}$ and $\bm{Z}$, and $P_{\bm{X}}(X)$ and $P_{\bm{Z}}(Z)$ are PMFs for $\bm{X}$ and $\bm{Z}$, respectively.\footnote{For continuous random variables, the summations are replaced with integrals, and PMFs are replaced with probability distribution functions (PDFs).} The logarithm base determines the information unit, which is \textit{nat} (natural base) in this paper.

An MI representation, derived from the Donsker-Varadhan definition of the Kullback-Leibler divergence \cite{Donsker:dvkl}, is 
\begin{equation}
    \label{eq:DV_representation}
    \mathcal{I}(\bm{X};\bm{Z}) {=} {\sup_{F: \Omega \rightarrow \mathbb{R}}} E_{P_{\bm{X,Z}}}[F(\bm{X},\bm{Z})] {-} \log (E_{ P_{\bm{X}}P_{\bm{Z}}} [e^{F(\bm{X},\bm{Z})}]).
\end{equation}
Here, $\Omega=\mathcal{X}\times\mathcal{Z}$ is the product space of the two random variables and the sub-index of the expectation shows the distribution over which the expectation is taken. The supremum is taken over all functions such that the expectations are finite. 

This optimization problem can be solved numerically using ML techniques, e.g., \cite{Belghazi:mine,Choi:remine,Song:smile} among others. For this, the sufficiently rich set of functions over which the search is performed is modeled with a network parameterized with a set of weights. Then, the optimal parameter that reaches the (local) maximum can be identified via stochastic gradient descent (SGD) techniques \cite{bottou-98x}. In addition, the two expectations are replaced with empirical averages over samples of a mini-batch that are drawn according to the joint distribution $P_{\bm{X,Z}}$ and the product of the marginal distributions $P_{\bm{X}}P_{\bm{Z}}$.

\subsection{Problem Description}

Our problem is to design an encoder $T_\Theta:\mathcal{X}\rightarrow\mathcal{Z}$ such that when applied to each sentence $X\in\mathcal{X}$, the encoded sentence satisfies certain theoretical information properties. Here, $\Theta$ represents the trainable parameters of the encoder{, and $L_i(X)$ and $S_j(X)$ represent the $i$-th relevant (public) feature and $j$-th irrelevant (sensitive) feature of $X$, respectively.} 
The architecture of the encoder is fixed, and its trainable parameters are optimized as follows:\vspace{-0.3cm}
\begin{equation}
    \begin{aligned}
        \max_\Theta & \, \gamma\mathcal{I}(T_\Theta(\mathbf{X});\mathbf{X}) + \sum_{i=1}^{N_l} \lambda_i \mathcal{I}(T_\Theta(\mathbf{X});L_i(\mathbf{X})) \\
        & - \sum_{j=1}^{N_s} \mu_j \mathcal{I}(T_\Theta(\mathbf{X});S_j(\mathbf{X})).
    \end{aligned}
\label{eq:main_opt}
\end{equation}
The parameters $\{\gamma,\lambda_1,\dots,\lambda_{N_l},\mu_1,\dots,\mu_{N_s}\}$  take non-negative values, and are tuned to enable a desired balance between competing goals of the optimization problem, {and the cardinality of $\bm{Z}$ determines the encoder's compression level.} {The distinction between (\ref{eq:main_opt}) and formulations found in prior research like \cite{beutel2017data} lies in the utilization of MI for training the encoder, as opposed to relying on the predictive performance of certain classifiers.}

Next, we describe the benefit of each term in the encoder optimization problem (\ref{eq:main_opt}). The first term is dedicated to information-theoretical compression to maximize the mutual information between a compressed sentence and its original representation. An information-theoretical compression stands above the classical methods, where simple distortion measures are minimized (e.g., mean square error after decompression). The second term is dedicated to utility by placing tunable attention on certain parts of the sentence content, making the embedding more representative of selected information. One benefit of such a scheme is that in scenarios where losing information is inevitable, it preserves all critical information. For example, through task-based compression of pilot scripts, the script execution process in airplane applications can be expedited without compromising safety. The last term is related to filtering out sensitive information from the encoded embedding either to make sure this sensitive information is not used by unauthorized users when data is publicly published (privacy), or to ensure biases are not incorporated into models trained using encoded representations (fairness). 

By properly setting the design parameters, a variety of interesting problems can be formed. For example, when $\lambda_i=\mu_j=0\;\forall i,j$, the problem is reduced to a \textit{task-agnostic} information-theoretical compression. When only $\lambda_i=0\;\forall i$, the problem is \textit{task-oriented} and is basically filtering out sensitive information in a   dataset. Finally, when $\gamma=0$, the \textit{task-oriented} problem is a privacy-utility trade-off where sentences in a dataset are encoded so that they can be used to train a model to identify public information (utility purpose). However, they are dramatically less useful for training a model to identify some sensitive information (for privacy purposes). In fact, the latter was discussed in \cite{Esfahanizadeh_2023} for the image domain with one public label and one private label.

\begin{figure*}[t]
  \centering
  \includegraphics[width=0.56\textwidth]{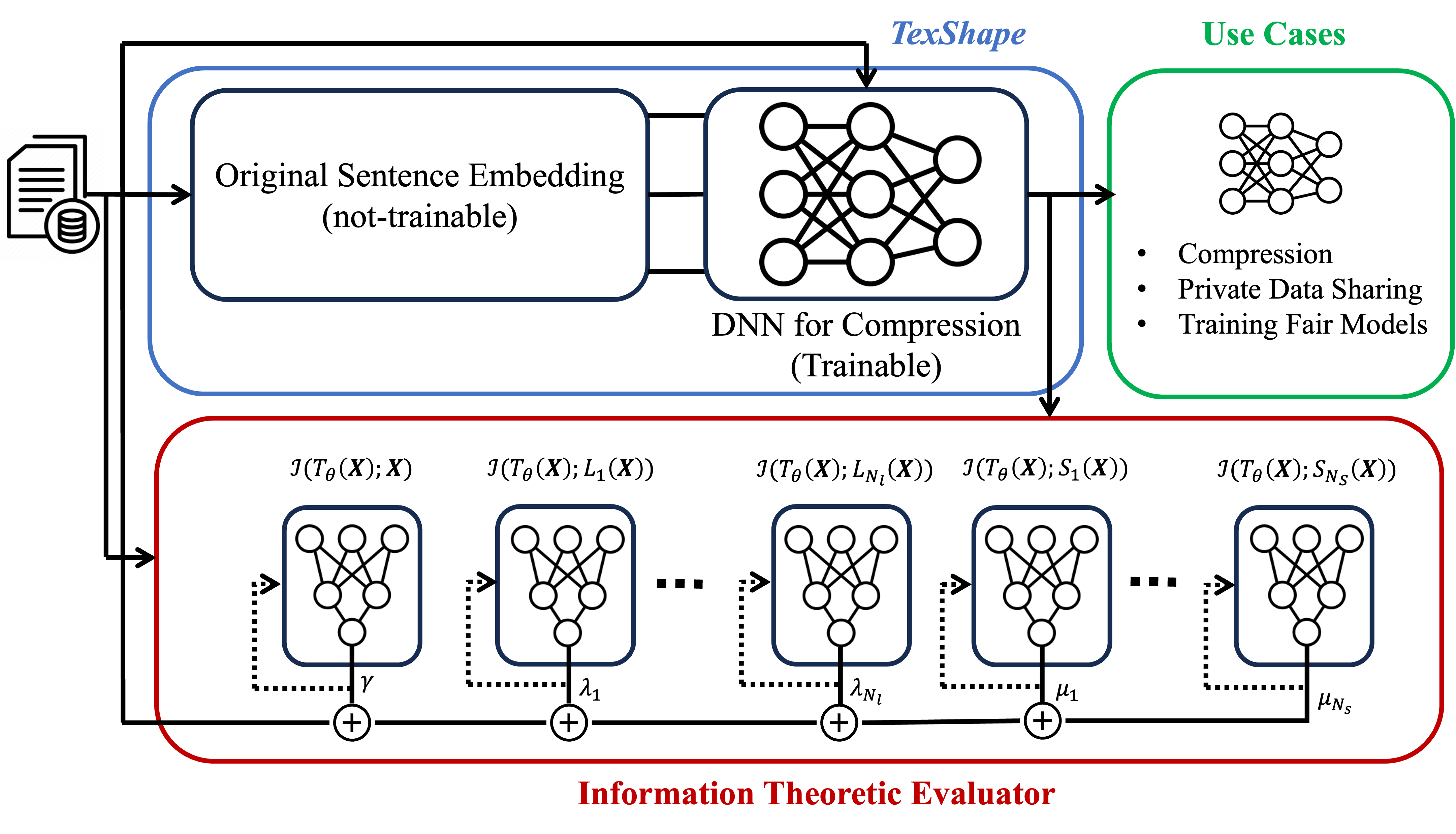}
  \caption{Architecture for designing and utilizing \texttt{TexShape}.}
  \label{pipeline}
\end{figure*}

\section{Designing \texttt{TexShape}}
In this section, we provide the architecture and the training procedure of \texttt{TexShape}, our novel information-theoretic sentence embedding framework.
The overall design strategy is illustrated in Fig.~\ref{pipeline}, which includes two main components: the sentence encoder \texttt{TexShape} which is highlighted with blue borders, and an information-theoretic evaluator that is used as a differentiable loss function to train \texttt{TexShape} and is highlighted with red borders. The trained encoder can be utilized on any dataset coming from the distribution that is used in the training stage, to customize its information-theoretic properties; for a variety of applications such as bandwidth-constrained transmission, private data-sharing, or unbiased model developments. The cost of using \texttt{TexShape} is equivalent to the cost of ML inference in NLP models.\vspace{0.1cm}

\noindent\textbf{Encoder.} The encoder consists of an off-the-shelf sentence embedding which sequentially receives tokens of a sentence and outputs a vector that represents the whole sentence. There are a variety of choices for this non-trainable part of the encoder such as \cite{devlin2018bert,song2020mpnet}. The sentence embedding is then fed into a neural network that projects the original sentence embedding into lower-dimensional embedding with certain information theoretical properties. This neural network is the trainable part of the encoder that is designed using an information theoretical evaluator, described next.\vspace{0.1cm}

\noindent\textbf{Information-Theoretic Evaluator.} 
This part consists of estimators for the MI terms that appear in the optimization problem (\ref{eq:main_opt}). MI is defined between two random variables $\bm{A}$ and $\bm{B}$, e.g., between a sentence embedding and its label or between two different embeddings for a sentence. We follow \cite{Belghazi:mine,Song:smile} and use the Donsker-Varadhan formulation of MI to empirically estimate it using samples drawn from the joint distribution. This formulation states that MI can be obtained by finding a function that maximizes an expression that depends only on the first moment of probabilistic terms; see (\ref{eq:DV_representation}). In this direction, the first moments can be empirically determined with the mean of the samples, known as the law of large numbers. In addition, the set of functions over which the search is performed can be modeled with a neural network, which is proven to model any well-behaved function with appropriate weights (Universal approximation theorem \cite{HornikUniversal1989}). Finally, optimization is performed using SGD, where at each step, a batch of samples is used in a public dataset to estimate expectations and their gradients. Then, the maximum converged value indicates the MI.


\noindent\textbf{Training.} To separate the iterations of training the sentence encoder from the iterations of MI estimators, we call the earlier an `epoch' and the latter an `iteration'. At each epoch, several MI values are estimated, each requiring many iterations. {To address the bias and instability inherent in mutual information estimation, we employ the techniques outlined in \cite{Choi:remine}.} As verified by our simulation results, the number of iterations is notably lower for higher epochs compared to the initial epochs. The encoder is initialized randomly for the first epoch, and at each epoch, its weights are updated to improve its information theoretical properties. MI values and their gradients can be iteratively estimated using batches of a public dataset from the intended distribution. After several epochs, the encoder can be isolated to be used by data owners.\vspace{0.1cm}

\noindent\textbf{Use Cases.} The encoded sentences can be used for a variety of task-agnostic or task-oriented applications. For example, when the encoder acts as a compression tool, the encoded data can be handled with less overhead in terms of storage, bandwidth, complexity, latency, etc. When the encoder acts as an encryption tool, certain sensitive features are removed, and the shared encoded data cannot be used by an adversary to infer sensitive individual properties of each sample. Finally, when the encoder acts as a fairness tool, certain sensitive information can be removed from training data to ensure downstream models do not incorporate biases in their decision-making. 

\section{Simulation Results}
In this section, we conduct our experiments on three public text datasets to optimize their representations according to different forms of the objective we define in (\ref{eq:main_opt}).

\noindent\textbf{Dataset~A: Stanford Sentiment Treebank (GLUE/SST2) \cite{socher-etal-2013-recursive}.}
It comprises sentences extracted from $67{,}349$ movie reviews, each annotated with a binary ``sentiment" label, being positive or negative, with $56{-}44$ ratio. We also generated a binary label per sample as ``sentence-length" with $50{-}50$ ratio, which indicates whether a sample contains more than eight tokens or not. The training data was then partitioned into a $90{:}10$ split, for  training and validation.\\ 
\noindent\textbf{Dataset~B: Corona-NLP \cite{covid-datasets}.}
It includes a set of Covid-related tweets, each annotated with a ``sentiment" label that has five different possibilities, i.e., extremely negative, negative, neutral, positive and extremely positive. During preprocessing, we made the sentiment label binary, indicating if the tweet is positive or negative, via excluding neutral samples and merging extremely negative with negative classes, and extremely positive with positive classes. As for the second label, we obtained per sample a binary ``location" label indicating if the author is from the United States or not. After preprocessing, the training and validation dataset was reduced to samples $14{,}335$ and $1{,}364$, where the location distribution was $56{-}44$ for training and $50{-}50$ for validation, while the sentiment distribution was $45{-}55$ for training $51{-}49$ for validation.\\
\noindent\textbf{Dataset~C: MultiNLI (Multi-Genre Natural Language Inference) \cite{N18-1101}.}
This dataset includes premise-hypothesis pairs of sentences, labeled entailment, contradiction, or neutral, specifying their logical relationship. In addition, each sample is labeled as government or telephone, based on its source and context. Following minimal preprocessing, $160{,}698$ training samples and $3{,}911$ validation samples are obtained from this dataset. The distributions of the two labels are $33{-}33{-}33$ and $48{-}52$ for the training data; and $36{-}33{-}31$ and $50{-}50$ for the validation data, respectively.

Next, we describe details of the \texttt{TexShape} encoder. The original model for sentence embedding (the non-trainable part of the encoder) is a pre-trained MPNet model \cite{song2020mpnet}, which is built upon Bidirectional Encoder Representations from Transformers (BERT) \cite{devlin2018bert}. The original sentence embedding gives a vector of size $768$ per sample. The trainable part is designed to be a DNN with $768$ input nodes, two hidden layers with $512$ and $256$ nodes in most cases, and an output layer whose cardinality is chosen based on the compression level, all utilizing the ReLU activation function. The training utilizes Adam optimizer with a learning rate $10^{-3}$, and has $20$ epochs. 

For an MI estimator, we used a DNN with the number of input nodes being a sum of the size of two random variables that their MI is estimated, and the number of output nodes being one. Each MI estimator has two intermediate layers with $64$ and $32$ nodes, respectively. The activation function for intermediate nodes is ReLU. The optimizer is Adam with the learning rate $10^{-4}$, and the number of optimization iterations is $2{,}000$. 
We ran our experiments with an NVIDIA RTX A5000 GPU with a memory of 24GB.


\subsection{\texttt{TexShape} to Address Privacy-Utility Trade-Off}
In this subsection, we utilize \textbf{Dataset~A} and \textbf{Dataset~B}. We aim to compress each datasets to filter out sensitive information while preserving critical information that is needed for certain downstream tasks. Thus, we choose the objective function to be a simplified form of (\ref{eq:main_opt}), where $\gamma=0$, $N_l=N_s=1$, $\lambda_1=1$ and $\mu_1=\mu$, resulting in
\begin{equation*}
        \max_\Theta  \, \mathcal{I}(T_\Theta(\mathbf{X});L(\mathbf{X})) - \mu \mathcal{I}(T_\Theta(\mathbf{X});S(\mathbf{X})).
\end{equation*}
Here, $L(\mathbf{X})$ denotes the ``sentiment" label for \textbf{Dataset~A} (SST-2) and ``location" label for \textbf{Dataset~B} (Corona-NLP), and $S(\mathbf{X})$ denotes the ``sentence-length" label for \textbf{Dataset~A} and ``sentiment" label for \textbf{Dataset~B}, respectively. Each MI is computed between a compressed representation with size $128$ and a binary label, and thus the network they use has $129$ input nodes. Each MI estimation iteration employs a batch size equal to the length of the full data set.



\begin{figure}[t]
    \centering
    \includegraphics[height=7cm, alt=Your Alt Text]{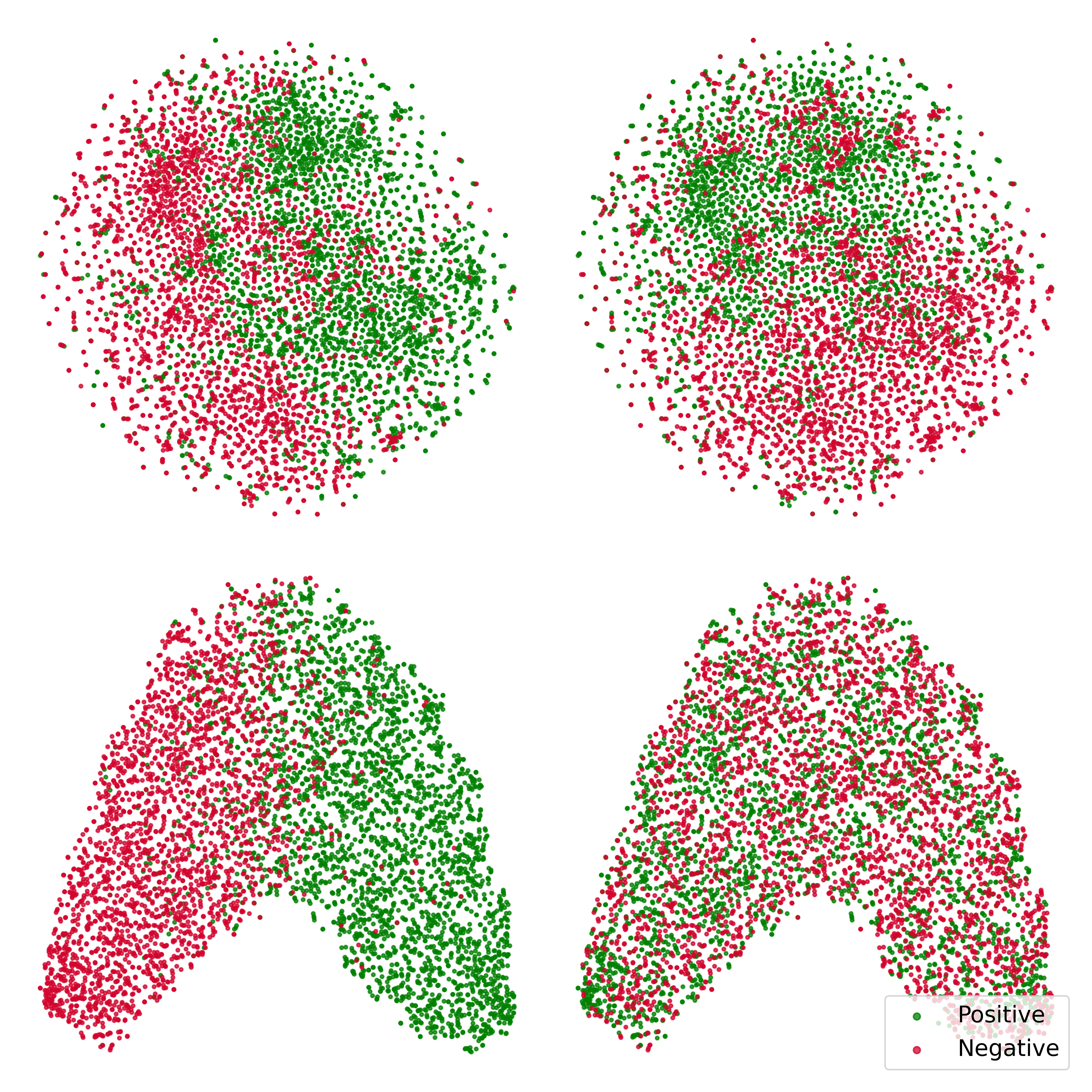}
    \caption{2D visualizations of sentences in \textit{Dataset~A}, colored based on their public label (left) and private label (right). The top panel is dedicated to the original embedding, and the bottom panel is dedicated to \texttt{TexShape} embedding.\vspace{-0.5cm}}
    \label{fig:sst2_tsne}
\end{figure}

Fig. \ref{fig:sst2_tsne} illustrates the T-SNE mapping obtained before and after employing the \texttt{TexShape} encoder trained with $\mu=0.3$ on \textbf{Dataset~A}. The illustrations show that the task-oriented encoder makes distinguishing between samples based on their private label visually impossible, while preserving the visual separability between samples based on their public labels. The estimated MIs between the embedding and the public label are $0.5051$ and $0.4739$ for original and \texttt{TexShape} embeddings, respectively. On the other hand, the estimated MIs between the embedding and the private label are $0.5552$ and $0.1334$ for the original and \texttt{TexShape} embeddings, respectively.

\begin{figure}[t]
  \centering
  \begin{tabular}{cc}
       \includegraphics[width=0.20\textwidth]{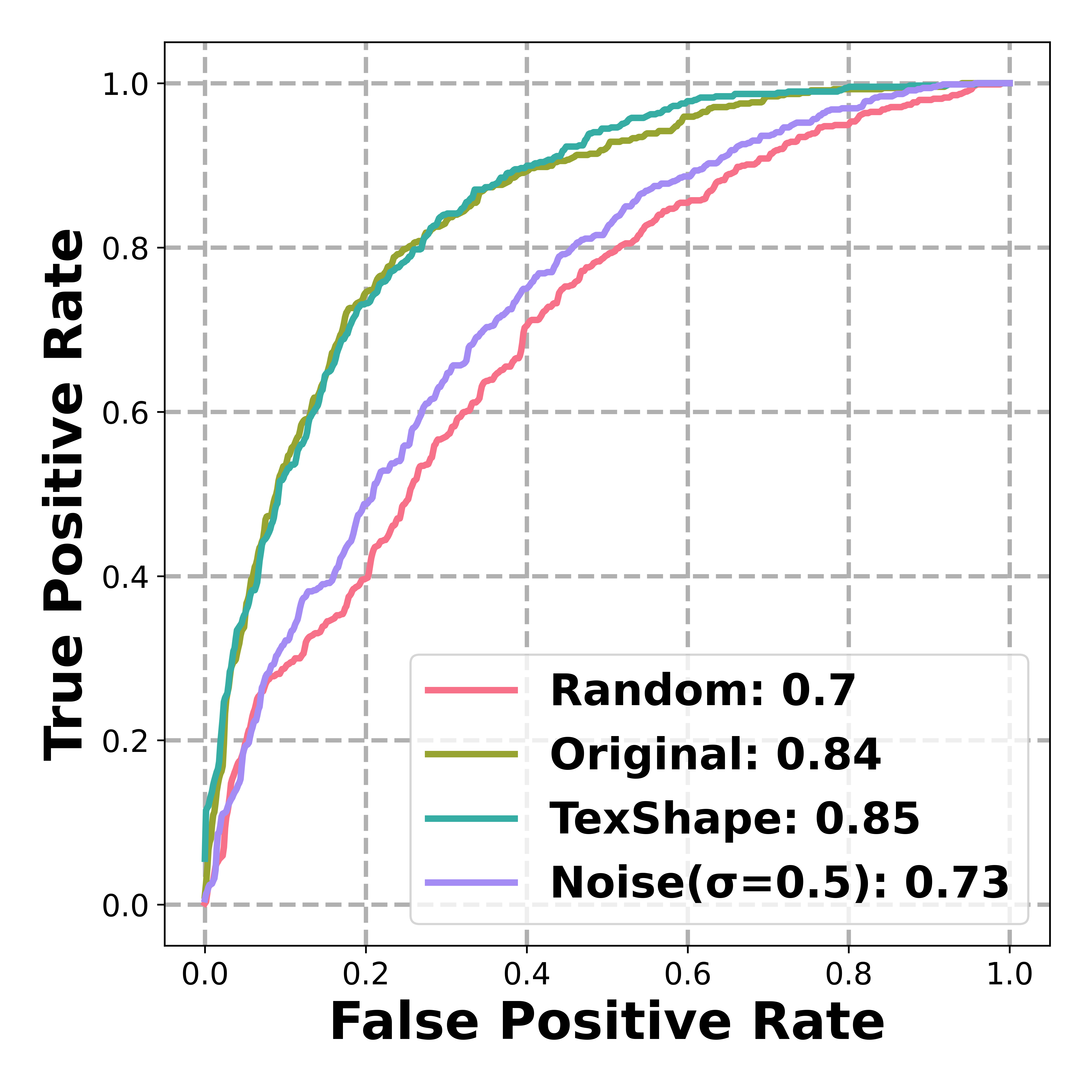}&
       \includegraphics[width=0.20\textwidth]{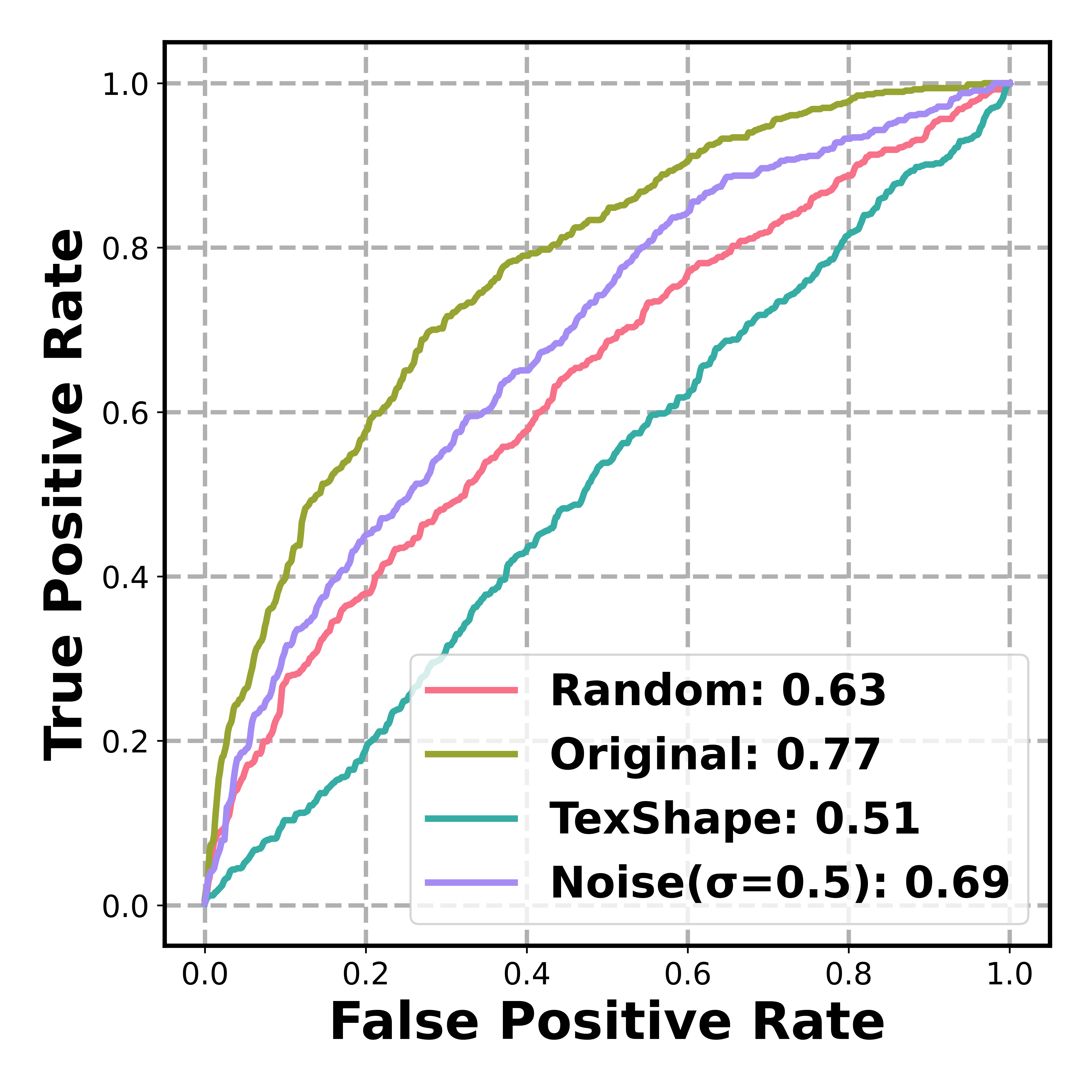}
  \end{tabular}
  \caption{{Validation ROC and AUROC for classifiers trained on \textbf{Dataset~B} for public task (left) and private task (right).}\vspace{-0.2cm}}
  \label{fig:ROC_curves}
\end{figure}

We also considered two classifiers trained to predict the public label and the private label of a compressed sample, respectively. We compared {four} different sentence embedding methods for \textbf{Dataset~B} as the training data: Original (when the information-theoretic compression part of the encoder is bypassed), Random (when weights of the trainable part of the encoder are chosen randomly), {Noisy (wherein the pixels of training data undergo independent distortion with Gaussian additive noise of mean $0$ and variance $\sigma^2$, aligned with the perspective of differential privacy),} and \texttt{TexShape} with $\mu=0.4$. Fig.~\ref{fig:ROC_curves} depicts the predictive performance of these classifiers on unseen data, in terms of their ROC curves, for public label and private label, respectively. As for the public label, the area under the ROC curve (AUROC) for \texttt{TexShape} even slightly improves the corresponding number for the original representation, suggesting that a compressed version can be more beneficial. As for private label, the AUROC for \texttt{TexShape} is around $0.51$, showing that this embedding can hardly be used by an adversary to identify sensitive information, even when similar public data exist. The performance of random embedding and {noisy embedding is worse} than \texttt{TexShape}, showing lower predictive utility and higher leakage of private information.


\subsection{\texttt{TexShape} for Information-Theoretic Compression}

In this subsection, we evaluate the compression capability of \texttt{TexShape} without emphasizing or filtering any specific feature. We utilize \textbf{Dataset~C}, and we aim to compress textual dataset by transforming the original embedding with size $768$ into a compressed embedding with lower dimensionality, while maximizing mutual information between the original embedding and the encoded embedding. Thus, the objective function derived from (\ref{eq:main_opt}) is simplified to $\max_\Theta  \, \mathcal{I}(T_\Theta(\bm{X});\bm{X})$. During \texttt{TexShape} training, only the premise part of each sample was utilized. 

For evaluation, we considered the generalization accuracy (ACC.) of two different classifiers, that is, predictor of the logical relationship between a pair of sentences (Label~1) and predictor of the source of a sentence (Label~2). 
The classifiers are trained on various embeddings of $\textbf{Dataset~{C}}$ and Table~\ref{tab:experiment2} summarizes the results, which manifests the power of our MI-based compression in preserving predictive utility in these two tasks. We would like to note that encoder training was agnostic to any of these features, and regardless, \texttt{TexShape} was successful in preserving higher information regarding each of these two features compared to untrained compression.

\begin{table}[t]
    \caption{Effect of information-theoretic compression on predictive utility of downstream classifiers.}
    \centering
    \begin{tabular}{|c|c|c|c|}
    \hline 
    {Embedding}&{Size}& Label 1 ACC. & Label 2 ACC. \\
    \hline
    Original Embedding & 768 & 69.3 & 99.4 \\
    Random Embedding & 128 & 51.1 & 90.8 \\
    \texttt{TexShape} Embedding & 128 & \textbf{61.0} & \textbf{95.3} \\
    Random Embedding & 64 & 35.9 & 82.4 \\

    \texttt{TexShape} Embedding & 64 & \textbf{57.3} & \textbf{95.5} \\

    



    \hline
    \end{tabular}\vspace{-0.2cm}
    \label{tab:experiment2}
\end{table}

\subsection{\texttt{TexShape} for Compression with Fairness Consideration}

In this subsection, we utilize \textbf{Dataset~A} and \textbf{Dataset~B}, and the focus is on keeping maximal information per sample on average, while filtering out certain information that can cause unwanted biases. Thus, the objective function derived from (\ref{eq:main_opt}) is $\max_\Theta \mathcal{I}(T_\Theta(\mathbf{X});\mathbf{X}) - 5 \mathcal{I}(T_\Theta(\mathbf{X});S(\mathbf{X}))$. Here, $S(X)$ represents the ``sentence-length" label for \textbf{Dataset~A} and ``sentiment" label for \textbf{Dataset~B}, respectively. We measure the MI between the embedding and the sensitive label as a measure of bias. Further, to evaluate the utility of the filtered dataset, we evaluated the performance of classifiers that are trained to predict ``sentiment" label for \textbf{Dataset~A} and ``location" label for \textbf{Dataset~B}. Table~\ref{tab:experiments3} summarizes the results: As we observe, the bias measure is dramatically lower in the outputs of \texttt{TexShape}, compared to the original datasets. This benefit comes at the cost of a small decrease in AUROC for a task that \texttt{TexShape} was agnostic to, compared to a similar setting where training was performed using the original dataset. {Our fairness method here is conceptually aligned with \cite{nature_dibiasing} where compression is used to de-bias the GloVe embedding. However, we do not restrict ourselves to a specific embedding.}

\begin{table}[h]
    \caption{Measure of utility and bias before and after \texttt{TexShape} encoding.}
    \centering
    \begin{tabular}{|c|c|c|c|c|c|}
    \hline
    \multirow{2}{5em}{Embedding} & \multirow{2}{1.4em}{Size} & \multicolumn{2}{|c|}{\textbf{Dataset~A}} & \multicolumn{2}{|c|}{\textbf{Dataset~B}} \\
    \cline{3-6}
    && AUROC & Bias & AUROC & Bias \\
    \hline

    Original & 768 & 0.9566 & 0.5832 & 0.8443 & 0.5552 \\ 
    \texttt{TexShape} & 128 & \textbf{0.9284} & \textbf{0.1939} & \textbf{0.7920} & \textbf{0.0711} \\
    \hline
    \end{tabular}







    \label{tab:experiments3}
\end{table}



\section{{Discussion and Future Work}}
In this paper, we have integrated fundamental concepts from information theory and machine learning and proposed a semantic approach for {data processing} in language models. We then highlighted the applications of our solution in areas such as lossy compression, privacy-preserving data sharing, and training unbiased models. 
{Our proposed design objective which is based on a weighted linear combination of data processing goals such as resource utilization, task-specific utility, privacy, and fairness, stands as a foundation for further exploration. For instance, expanding the linear combination to more general forms is promising in achieving a better balance among competing goals. Furthermore, optimizing the weights as hyperparameters, informed by theory, can bridge a link between design choices and desired performance metrics. These avenues are key directions for our future studies.}

\section*{Acknowledgment}
This research is supported in part by a grant from the CII (Construction
Industry Institute) - OS2 program, MIT Portugal Program, and a grant by JMA Wireless.

\newpage
\bibliographystyle{IEEEtran}
\bibliography{references}

\end{document}